\documentclass[lettersize,journal]{IEEEtran}
\usepackage{amsmath,amsfonts}
\usepackage{algorithmic}
\usepackage{algorithm}
\usepackage{array}
\usepackage[caption=false,font=normalsize,labelfont=sf,textfont=sf]{subfig}
\usepackage{textcomp}
\usepackage{stfloats}
\usepackage{url}
\usepackage{verbatim}
\usepackage{graphicx}
\usepackage{multirow}
\usepackage{cite}
\usepackage{pifont}
\usepackage{xcolor} 
\usepackage[table,xcdraw]{xcolor} 
\usepackage{hyperref} 

\newcommand{\cmark}{\ding{51}} 
\newcommand{\xmark}{\ding{55}} 

\DeclareRobustCommand{\IEEEauthorrefmark}[1]{\smash{\textsuperscript{\footnotesize #1}}}

\begin{document}

\title{Efficient-Husformer: Efficient Multimodal Transformer Hyperparameter Optimization for Stress and Cognitive Loads}

\author{ 
{Merey Orazaly}\IEEEauthorrefmark{1}\textsuperscript{\textdagger},
{Fariza Temirkhanova}\IEEEauthorrefmark{1}, and
{Jurn-Gyu Park}\IEEEauthorrefmark{1}\textsuperscript{*}


%
\thanks{This manuscript is currently under review.
}

\thanks{
}

\thanks{\textsuperscript{1}The School of Engineering and Digital Sciences, Nazarbayev University, Astana, Kazakhstan (\textsuperscript{*}corresponding author: jurn.park@nu.edu.kz). 
\textsuperscript{\textdagger}Work started as a MS thesis at Nazarbayev University.
}}

\markboth{
}%
{Shell \MakeLowercase{\textit{et al.}}: A Sample Article Using IEEEtran.cls for IEEE Journals}


\maketitle

{\color{black}
\begin{abstract}
Transformer-based models have gained considerable attention in the field of physiological signal analysis. They leverage long-range dependencies and complex patterns in temporal signals, allowing them to achieve performance superior to traditional RNN and CNN models. However, they require high computational intensity and memory demands. In this work, we present Efficient-Husformer, a novel Transformer-based architecture developed with hyperparameter optimization (HPO) for multi-class stress detection across two multimodal physiological datasets (WESAD and CogLoad). The main contributions of this work are: (1) the design of a structured search space, targeting effective hyperparameter optimization; (2) a comprehensive ablation study evaluating the impact of architectural decisions; (3) consistent performance improvements over the original Husformer, with the best configuration achieving an accuracy of 88.41 and 92.61 (improvements of 13.83\% and 6.98\%) on WESAD and CogLoad datasets, respectively. 
%
The best-performing configuration is achieved with the $(L+d_m)$ or $(L+FFN)$ modality combinations, using a single layer, 3 attention heads, a model dimension of 18/30, and FFN dimension of 120/30, resulting in a compact model with only \textasciitilde30k parameters. 

%
\end{abstract}

}

\begin{IEEEkeywords}
Hyperparameter Optimization (HPO), Multimodal Transformer, Attention Mechanisms, Wearable Sensor Data, Physiological Signal Processing, Stress Detection, Cognitive Load
\end{IEEEkeywords} 


\section{Introduction}
{\color{black}
Transformer-based architectures have gained significant attention across a broad range of research domains in the processing of physiological signals. It has the capacity to capture long-range dependencies and extract features from complex, high-dimensional data.
Since conventional Recurrent Neural Networks (RNN) and Convolutional Neural Networks (CNN) \cite{Keren2016ConvolutionalRA} process sequences in a local manner and are not capable of processing sequences in their entirety, self-attention-based Transformer models are therefore well-suited for performing stress detection tasks. 
Transformer-based models yield considerable improvements for healthcare applications such as ECG classification \cite{behinae}, 
emotion recognition \cite{Vu2023MultiscaleTN}, and monitoring based on wearables \cite{Lange2023PrivacyPreservingSD}. The recent literature has dealt with Transformer extensions like Time Series Transformers (TST) and Physiological Signal Transformers (PST) \cite{yang2024decompose} \cite{vazquezrodriguez2022}, demonstrating that these models can effectively manage multimodal biosignals and enhance the accuracy of classification. 

However, a big challenge is the optimization of Transformer architectures toward real-time and resource-efficient implementation. Although the Transformers were successful in physiological signal analysis, the challenge that exists for the design of efficient Transformer models towards the multi-class stress detection is linked to their immense computational load and memory requirements coupled with trade-offs between accuracy and efficiency. 
Standard Transformer architectures need a considerable amount of computational resources; therefore, they are impractical for performing real-time operations and resource-constrained platforms such as mobile and wearable devices \cite{limittransf}. 
In the case of multi-class stress detection, the model has to preserve generalization properties over various states of physiological signals while also prioritizing low latency and energy consumption \cite{electronics14040687}.

Mostly, existing approaches primarily concentrate on accuracy, ignoring the practicalities of Transformer deployment in real-world contexts with hardware limitations. Addressing those problems requires systematic research on hyperparameter optimization (HPO) for Transformer models, specifically focusing on layers ($L$), attention heads ($H$), dimension ($d_m$) and FFN dimension ($FFN$) in order to provide a practical and scalable framework for stress detection tasks. 

In this paper, we propose a Husformer-based~\cite{husformer} optimized model, a highly efficient Transformer-based model applied to multi-class stress detection using the WESAD \cite{wesad} and CogLoad \cite{cogload} datasets. 
We introduce an optimization strategy that adjusts systematically the number of parameters to reach an optimal trade-off between accuracy and efficiency.

}{\color{black}
The paper specifically highlights the following contributions: 



\begin{itemize}
    \item Propose a hyperparameter optimization strategy that employs a constrained local search space to enable efficient tuning of Transformer models (e.g., Husformer) for performance improvements in resource-constrained platforms, using small-scale affective computing and cognitive load datasets of WESAD and CogLoad.

    \item Perform a systematic ablation study to identify combinations of the most critical components that contribute to the model's effectiveness. 
    


    
    \item The Efficient-Husformer achieves the accuracies of 88.41 and 92.61 on WESAD and CogLoad, respectively, corresponding to substantial improvements of 13.83\% and 6.98\% over the original Husformer results, with a dramatically reduced number of parameters. 
    
    %
    Our open-source codes are on GitHub\footnote{\url{https://github.com/Merey1508/Efficient-Husformer}}.
    %
    %
    %

\end{itemize}

The rest of the paper is organized as follows: Section II presents the background, motivation, and summary of related works. The search space and optimization strategy are proposed in Section III. The descriptions of experimental settings, datasets, evaluation, and quantitative metrics are outlined in Section IV. Section V contains summaries of the quantitative assessment, ablation study, and results and analysis. The discussion and future work are presented in Section VI, and the paper is concluded in Section VII. 

{\color{black}
\section{Motivation and related work}

\subsection{Background}


The Husformer is defined by three major components, which are illustrated in Figure \ref{figure:architecture}: Before Cross-Modal Transformer, Cross-Modal Transformer, and Self-Attention Transformer for final prediction.}

\begin{figure*}
\centering
\includegraphics[width=1\linewidth]{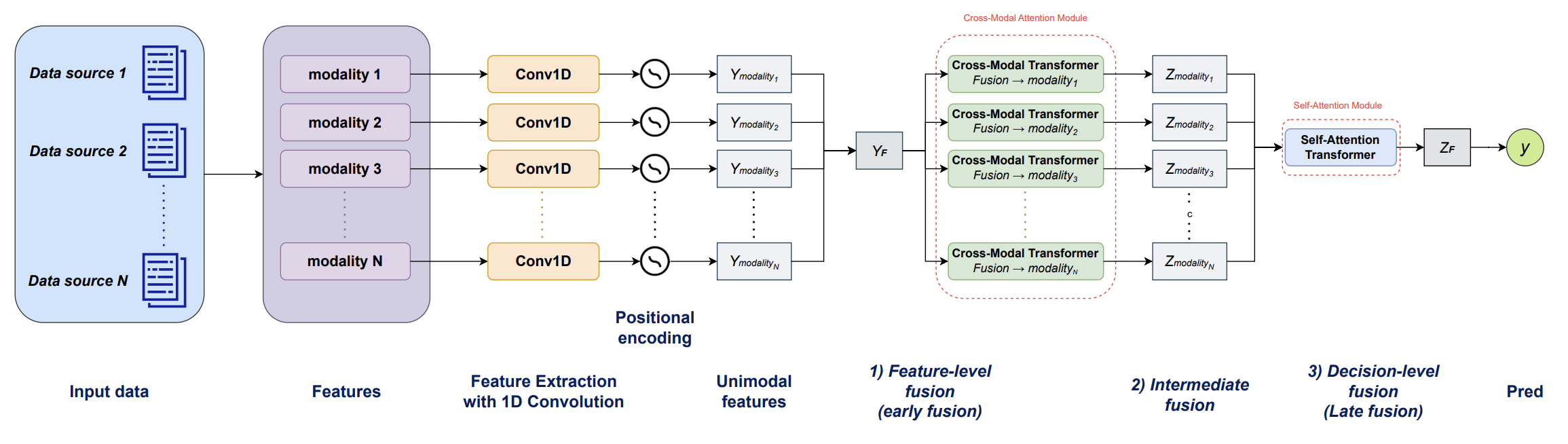}
\caption{{\color{black}Efficient-Husformer Architecture (Deployed from \cite{husformer} and Modified).}
}
\label{figure:architecture}
\end{figure*}

{\color{black}
\subsubsection{Before Cross-Modal}
In the first phase, multimodal physiological signals (e.g., GSR, BVP, EMG, ECG, RESP) are treated as discrete input streams (modalities). Each modality is processed separately through a 1D convolutional layer (Conv1D) that highlights spatial features of that signal type. 
After convolution, positional encoding is added to each unimodal feature vector to reflect the order in which features are extracted – which is essential for time series data. The end result is a sequence of unimodal feature vectors, one per source, in a fixed-dimensional space. 
The unimodal features are then concatenated together to form a joint representation, labeled $Y_F$. This is the early fusion or feature-level fusion, where features from all modalities are taken together before reaching the Transformer layers.


\subsubsection{Cross-Modal Transformer}

In this core module, it enriches each unimodal feature by providing contextual information from the other modalities using a Cross-Modal Transformer. The cross-modal Transformer consists of $U$ stacked encoder layers, where each layer consists of 
a Multi-Head Cross-Modal Attention Block and a Position-Wise Feed-Forward Network (FFN).
Given the Transformer architecture, all components are wrapped in residual connections and layer normalization. Provided a unimodal target feature sequence $Y_{M_i}$ and a shared fused representation $Y_F$, the function to transform between the encoder layers is defined as follows: 

\begin{align}
Z^{[0]}_F &= Y_F \\
Z^{[0]}_{M_i} &= Y_{M_i} \\
\hat{Z}^{[u]}_{M_i} &= \text{CM}^{[u]}_{\text{Mul}}\left( L(Z^{[0]}_F), L(Z^{[u-1]}_{M_i}) \right) \\
\dot{Z}^{[u]}_{M_i} &= \hat{Z}^{[u]}_{M_i} + L(Z^{[u-1]}_{M_i}) \\
Z^{[u]}_{M_i} &= F_{\theta} \left( L(\dot{Z}^{[u]}_{M_i}) \right) + L(\dot{Z}^{[u]}_{M_i})
\end{align}

where $u \in \{1, \ldots, U\}$  is the encoder layer index, $L(\cdot)$ is layer normalization, $F_\theta(\cdot)$ is the position-wise feed-forward network, and $\text{CM}^{[u]}_{\text{Mul}}$ is the cross-modal attention function used in the transformation step based on the fact that it uses $Z^{[0]}_F$ as keys and values, and $Z^{[u-1]}_{M_i}$ as query.

\subsubsection{Self-Attention Transformer}
The last stage of the Husformer architecture processes decision-level fusion with a Self-Attention Transformer. After cross-modal encoding, the updated modality-specific representations $\{Z^{[U]}_{M_i}\}$ are merged and passed into a Transformer module with $L$ layers, where each layer has $H$ attention heads along with $d_{ffn}$-sized feed-forward blocks. 
In this Transformer, global dependencies across all modality embeddings will be captured through the multiple layers, allowing the model to hone in on the joint representation $Z_f$ by attending to complementary characteristics across modalities. The result is a context-aware, high-level feature vector used in a prediction head to generate the model output: 
\[
y = \text{Pred}(Z_f)
\]
This way of allowing flexible interaction across the modality-specific signals help the model make better and more robust predictions by virtue of the temporal and contextual patterns learned across the other stages of the pipeline.
}

\begin{figure}[t]
\centering
\includegraphics[width=0.9\linewidth]{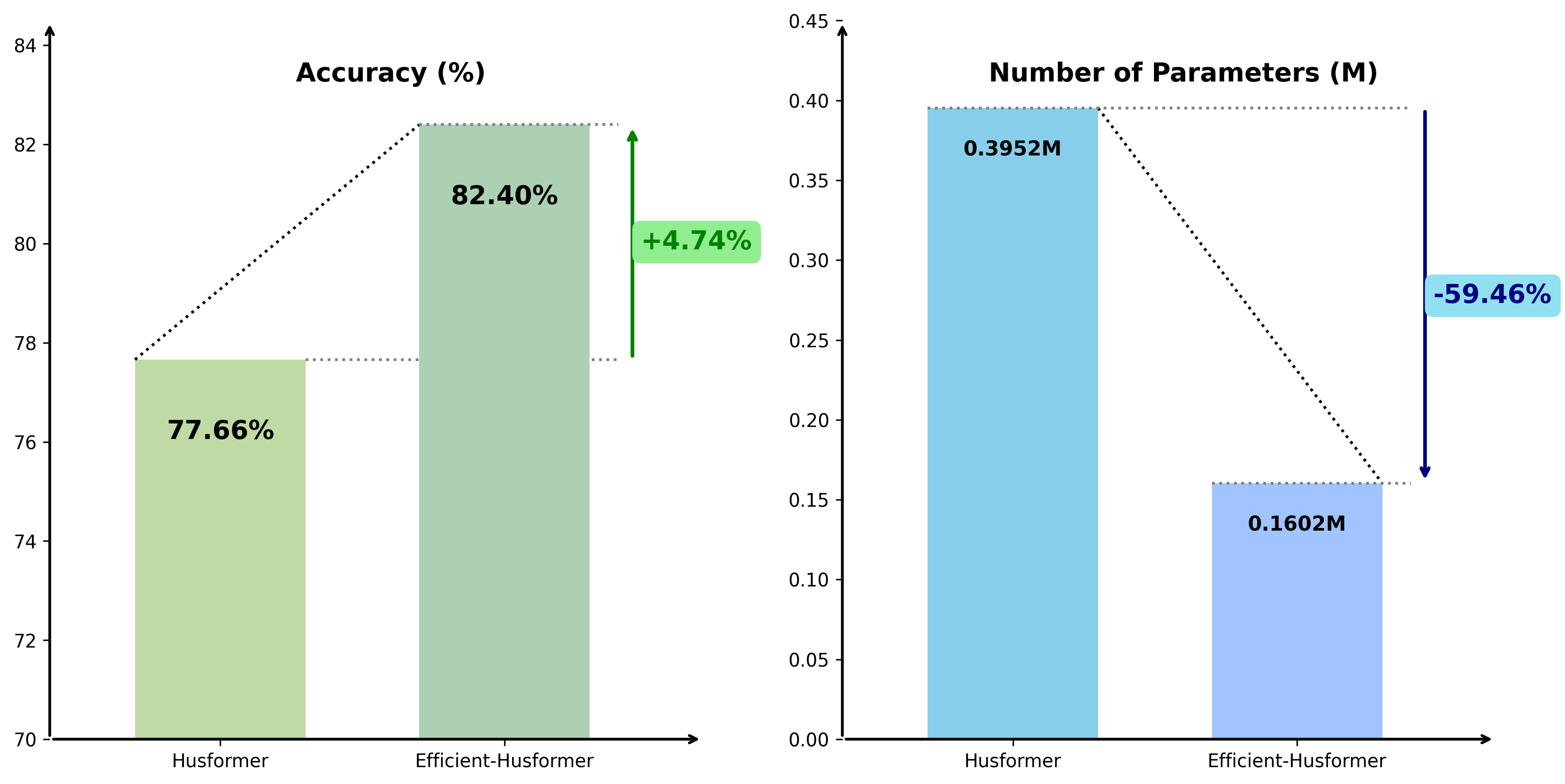}
\caption{{\color{black}Motivating Example: Comparison of Original ($L$=5) and an Efficient layer ($L=3$) in the other default configurations: Heads ($H=3$), Dimension size ($d\_m=30$), and Feed-Forward Network size ($FFN=120$)}
} 
\label{figure:motivating example}
\end{figure}

{\color{black}
\subsection{Motivation}


The standard Transformer architectures are traditionally built with fixed hyperparameters such as \textbf{number of layers ($L$)}, \textbf{attention heads ($H$)}, \textbf{model dimension size ($d_m$)}, and \textbf{FFN dimension size ($FFN$)}, which may lead to sub-optimal trade-offs between accuracy and efficiency \cite{chitty2022}. 
In this study, we work toward optimizing the Transformer model by a systematic tuning of these hyperparameters while still maintaining a competitive performance. We have defined the configurations of the Transformer models as: 
\begin{equation}
    \text{Transformers Model Variants} = f(L, H, d_m, FFN)
    \label{perf}
\end{equation}



\noindent{\textbf{Motivating example}: 
As a motivating example, we consider the baseline Husformer model, varying only a single hyperparameter — the number of layers $L$, which is one of the most effective in terms of performance — while keeping all other hyperparameters fixed at their default values. 
Surprisingly, our findings in Figure \ref{figure:motivating example} show a 4.74\% increase in accuracy, accompanied by a substantial reduction of 59.46\% in the number of parameters,  highlighting improvements in both accuracy and computational efficiency. 
%
This example provides strong empirical opportunities for re-evaluating state-of-the-art Transformer models and identifying efficient configurations better suited for multimodal datasets.}
}

\subsection{Related Work}
\begin{table*}[b]
\centering
\caption{Summary of Related Works}\vspace{0.5em}
\renewcommand{\arraystretch}{1.2}
\setlength{\tabcolsep}{3pt} 
\begin{tabular}{|>{\centering\arraybackslash}p{0.13\textwidth}|
                >{\centering\arraybackslash}p{0.08\textwidth}|
                >{\centering\arraybackslash}p{0.1\textwidth}|
                >{\centering\arraybackslash}p{0.09\textwidth}|
                >{\centering\arraybackslash}p{0.10\textwidth}|
                >{\centering\arraybackslash}p{0.15\textwidth}|
                >{\centering\arraybackslash}p{0.25\textwidth}|}
\hline
\textbf{Study} & \textbf{Multi-class} & \textbf{Dataset} & \textbf{CV} & \textbf{Modalities} & \textbf{Auto. Extracted Features} & \textbf{Algorithms (Accuracy)} \\ 
\hline\hline
\multicolumn{7}{|c|}{\textbf{Classical ML Models}} \\ 
\hline
Schmidt et al. (2018) \cite{wesad} & \cmark & WESAD & \xmark & 7 & \xmark  & DT, RF, AB, LDA, KNN (74.20\%–87.74\%) \\ 
\hline
Aqajari et al. (2021) \cite{PYEDA2021} & \xmark & WESAD & LOOCV & 1 & \cmark  & KNN, RF, SVM, NB (85\%–91.60\%) \\ 
\hline
Bobade et al. (2020) \cite{bobade} & \cmark & WESAD & K-fold (k=5) & 7 & \xmark  & DT, RF, AB, LDA, KNN, SVM, ANN (87.59\%–95.21\%) \\ 
\hline
Su et al. (2022) \cite{suet} & \cmark & WESAD & K-fold (k=10) & 3 & \xmark & RF, LR, SVM, FNN (84.62\%) \\ 
\hline\hline
\multicolumn{7}{|c|}{\textbf{DL Models}} \\ 
\hline
Behinaein et al. (2021) \cite{behinae} & \xmark & ECG & \xmark & 1 & \cmark  & Transformers (80.4\%) \\ 
\hline
Yao et al. (2021) \cite{yao-etal-2021-muser} & \cmark & Custom & K-fold (k=5) & 2 & \cmark & MUSER Transformer (84.2\%) \\ 
\hline
Ziaratnia et al. (2023) \cite{ziarat} & \cmark & Custom & \xmark & 1 & \cmark & CCT-LSTM (83.2\%) \\ 
\hline
Husformer. (2024) \cite{husformer} & \cmark & WESAD & \xmark & 6 & \cmark & Transformers (78.68\%) \\ 
\hline
\end{tabular}
\label{related}
\end{table*}


Over the past years, a number of Machine Learning (ML) and Deep Learning (DL) models have been widely used in the domain of stress detection, considering different algorithms to analyze physiological and behavioral data.
As shown in Table \ref{related}, we summarize our related works with the three categories of classical ML (non-DL), general DL (non-Transformers), and Transformer-based models, considering the modalities used, whether features were automatically extracted, cross-validation methods, and algorithms applied alongside their reported accuracies in classification. 
%

\subsubsection{Classical ML (non-DL) models} 
A notable study by Schmidt et al. \cite{wesad} discusses the use of multimodal data for affect detection using multiple sensor modalities, such as Accelerometer (ACC), Blood Volume Pulse (BVP), Electrodermal Activity (EDA), Skin Temperature  (TEMP), Respiration (RESP), Electromyogram (EMG), and Electrocardiogram (ECG), along with a wide variety of classifiers such as Decision Tree (DT), Random Forest (RF), Ada Boost (AB), Linear Discriminant Analysis (LDA), and K-Nearest Neighbors (KNN), with accuracy rates between 74.20\% and 87.74\% using the leave-one-subject-out (LOSO) method.  
{\color{black}Another study by Aqajari et al. \cite{PYEDA2021} develops a model that uses features of pyEDA, a Python toolkit for EDA, for pre-processing and statistical and automatic feature extraction. In the binary classification task for stress recognition, their model achieves an accuracy of 79.71\% on the WESAD dataset, surpassing previous approaches.}

%

Bobade et al. \cite{bobade} incorporate Support Vector Machine (SVM) and Artificial Neural Network (ANN) models, yielding higher accuracy values of 87.59-95.21\%. The performance improves from increasing the number of classifiers, allowing better handling of the complexity and variability of the data between subjects.
In the study of Su et al. \cite{suet}, four ML algorithms were developed for college students' models, by developing stress prediction models: RF, Logistic Regression (LR), SVM, and Feed-Forward Neural Network. The RF model resulted in the best predictive capability for stress levels, with the highest performance among all models, by reaching an accuracy of 84.62\%, specificity of 96.35\%, and AUC and F1 of 82\%. Their work highlighted the importance of specific modalities in achieving good performance while maintaining the complexity of the model as low as possible. 

\subsubsection{DL models} 
Attention mechanisms have further enhanced the extraction of relevant features by assigning varying importance to different data segments \cite{vaswani}, which is crucial in identifying critical moments of stress-related physiological changes. 
Using this approach, Behinaein et al. \cite{behinae} presented only ECG data, using Transformers for feature extraction, and reported an accuracy of 80.4\% by employing the LOSO cross-validation technique.
Yao et al. \cite{yao-etal-2021-muser} proposed MUSER, a Transformer-based model whose performance in the task of detecting stress was facilitated by emotion recognition as an auxiliary task. Consequently, MUSER relied on the interdependence between the two variables (stress and emotion) and achieved 84.2\% accuracy results in the Multimodal Stressed Emotion-MuSE dataset, an indication of the multi-task learning benefit in performing affective computing. These studies further point out that deep learning models improve the process of stress detection by using multimodal data and sophisticated architectures. 
Ziaratnia et al. \cite{ziarat} proposed a novel method on remote video-based stress estimation using a Convolutional Channel-wise Transformer combined with Long Short-Term Memory (CCT-LSTM). Their approach provided better performance on the spatial and temporal features extracted from facial cues, yielding an accuracy of 83.2\% and an F1 score of 83.4\%. 

{\color{black}
Finally, Wang et al. \cite{husformer}, an end-to-end multimodal Transformer named Husformer (as our reference model), which learns from multimodal data streams. Utilizing cross-modal attention and self-attention transformers, it effectively identifies relevant contextual information and captures long-term temporal dependencies in human affective states. Husformer consistently outperforms the state-of-the-art multimodal baselines and single-modality methods, including datasets without further feature engineering. Specifically, on WESAD and CogLoad, it achieved accuracy results of 78.68\% and 74.06\%, respectively, revealing superior adaptivity and capacity for meaningful and effective pattern recognition. }

{\color{black}
\section{Methodology}



This section outlines the essential search space and optimization strategy that conceptualize and develop efficient Transformer models applied to different datasets, 
%
searching for efficient Transformer architectures that maximize multimodal learning without sacrificing task-specific performance. 


\begin{figure}
\centering
\includegraphics[width=0.9\linewidth]{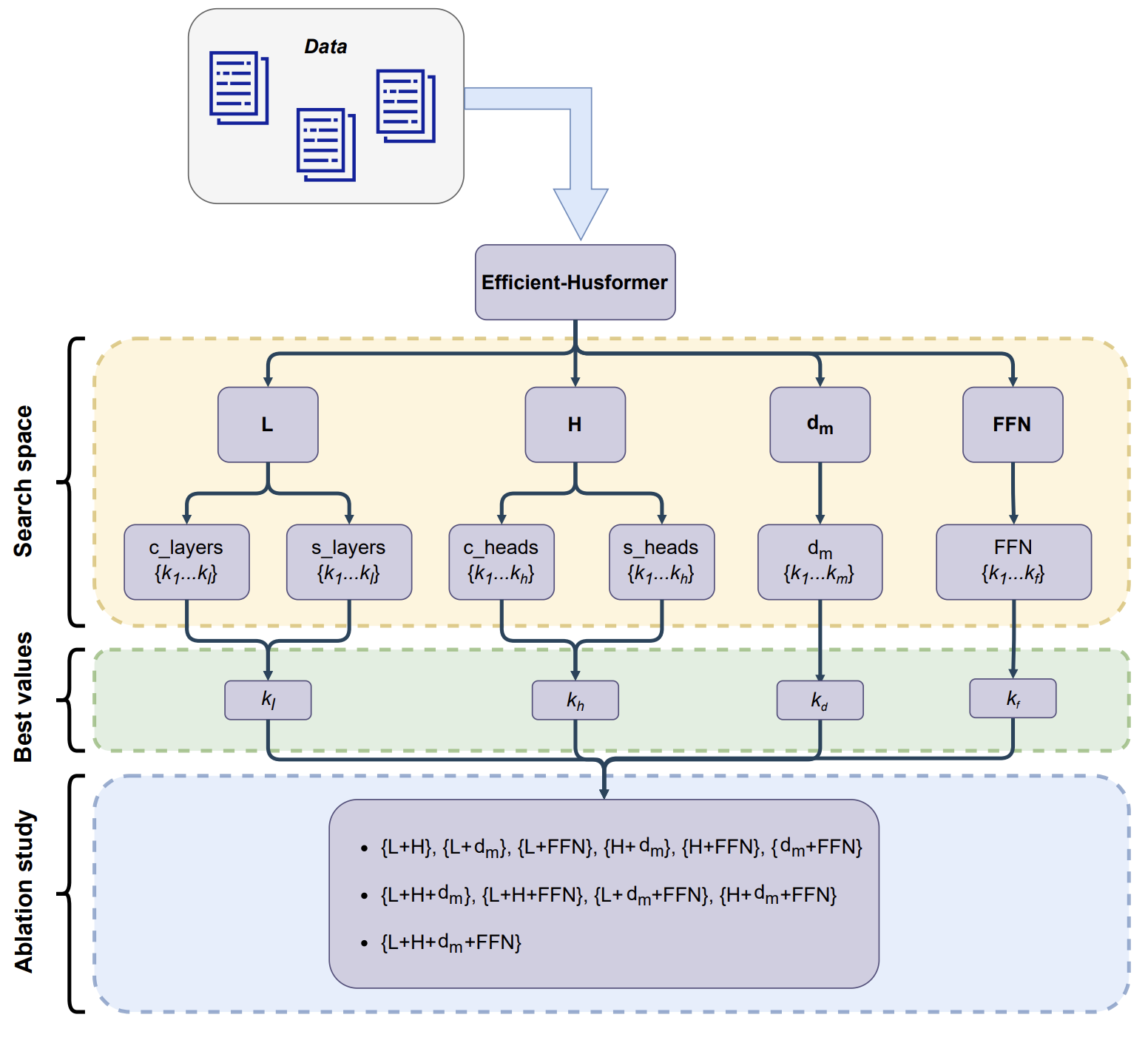}
\caption{{\color{black}Methodology Overview}}
\label{figure:optimization}
\end{figure}

} 

{\color{black}
Taking into account the multimodal characterization of their physiological signals, we further introduce a decoupled optimization strategy in which the cross-modal Transformer ($L_{cm}$, $H_{cm}$) 
and the self-attention Transformer ($L_{sa}$, $H_{sa}$) for modality fusion are separately optimized:  


\begin{equation}
    f(L_{cm},H_{cm}) + f(L_{sa}, H_{sa}) + f(d_m) + f(FFN)
    \label{perf}
\end{equation}

Therefore, this parametric configuration allows 
independent optimizations for both modality fusion and classification functionalities for improved efficiency and interpretability.
To effectively navigate this optimization space and strategy, we aim to address the following three questions based on the three steps (Figure~\ref{figure:optimization}):

\begin{itemize}
  \item {\it How can a structured search space combined with local optimization strategies guided by priority assumptions improve the efficiency and effectiveness of hyperparameter optimization in Transformer-based models?}
  
  \item {\it Which architectural and training components identified from a systematic ablation study are the most important to the performance of the Transformer model?
  
  \item {\it To what extent do the proposed Efficient-Husformer models outperform the original Husformer in terms of classification accuracy, and does its open-source implementation facilitate reproducibility?}}
\end{itemize}

\noindent{With these three questions, we establish a comprehensive optimization framework that balances performance, resource efficiency, and deployment feasibility. 
The study characterizes the influence of components of the Transformer on tasks for stress detection and offers further guidance on efficient design for Transformer models.}

\begin{table}[t]
    \centering
    \caption{Search Space 
    }
    
    \label{tab:optimization_space}
        \begin{tabular}{|l|l|c|}
        \hline
        \rowcolor[HTML]{C0C0C0}
        \multicolumn{3}{|c|}{\textbf{Model Size}} \\
        \hline
        \multirow{3}{*}{Attention} & Layers & $\{k_1, k_2, \dots, k_l\}$ \\
        \cline{2-3}
        & Heads & $\{k_1, k_2, \dots, k_h\}$ \\
        \cline{2-3}
        & Model Dimension & $\{k_1, k_2, \dots, k_m\}$ \\
        \hline
        FFN & FFN Size & $\{k_1, k_2, \dots, k_f\}$ \\
        \hline
    \end{tabular}
\end{table}

\subsection{Search Space}


For the optimization space, as shown in Table~\ref{tab:optimization_space}, we simultaneously focus on key architectural hyperparameters to balance computational efficiency and representational power. 
For each hyperparameter, we access a smaller but still expressive range of values from the baseline configurations. The constraint of hyperparameters objectively allows us to systematically and reproducibly optimize, without generating exhaustive enumerations of parameters. 

More importantly, we believe the parameter space captures the optimal configurations for maximizing efficiency (latency, memory, and energy consumption). This parameter space also enables a variety of architectural variants that are still expressive enough to model multimodal dynamics (e.g., stress detection) across several datasets.

\subsection{Optimization Strategy}

As shown in Figure~\ref{figure:optimization}, the optimization process follows a rigorous three-step process:

\begin{enumerate}
    \item \textbf{Defining the search space:} In the first step, we establish a search space that is bounded and meaningful for the model architecture and the data specifics. Each hyperparameter is given a finite set of discrete values.
    
    \item \textbf{Identifying the best configurations:} In the second step, we have candidate combinations that are marked according to a trade-off between predictive performance and computational cost. Candidate models with the best performance in this stage are selected for the next tuning phase of the study. 
    
    \item \textbf{Ablation study:} In the final step of the process, we conduct a systematic ablation study to fully understand the contribution of the individual and combined contributions of the core components of the structure. We conduct experiments where we vary or eliminate one component at a time to see the contribution of the core component. We also test all combinations of pairs (e.g., ($L$ + $H$), ($H$ + $d_m$)), triples (e.g., ($L$ + $H$ + $FFN$)), and the full combination (all four modules).
\end{enumerate}

This exhaustive approach allows us to examine the individual contributions of the components through additive and interaction effects.
Importantly, the search space is constructed in a way that maintains architectural constraints; for example, $d_{m}$ values must be divisible by $H$ for multi-head attention components. This constraint-based design minimizes the space to only those computationally valid parameter settings, while reducing the chance of invalid model generations occurring. Therefore, only operationally sensible and computationally feasible parameter space for optimization is used. 

Specifically, the model dimension ($d_{m}$) serves a dual purpose: it defines the size of the token embedding dimensions and simultaneously acts as the projection space for the attention mechanism; therefore, the feed-forward network (FFN) is defined by two linear layers, and it is tied to the model dimensions through a scale factor, $\alpha$, resulting in the hidden size of FFN as ($\alpha$ $\times$ $d\_m$).
%
The coupling of these metrics indicates that tuning the FFN, along with the model dimension, is crucial in maintaining consistency and allowing for valid projections across layers. On the other hand, the number of attention heads and the number of Transformer layers on both self-attention and cross-modal branches can be considered relatively independent. 


In order to progressively evaluate the relative importance of each hyperparameter in the Efficient-Husformer architecture, we conduct a simple yet effective hyperparameter optimization (i.e., parameter isolation) approach. 
Through this approach, we measure the performance difference for a hyperparameter by varying its value while keeping all other parameters at their default settings \cite{HPI}. This strategy allows for the direct assessment of individual hyperparameter contributions to model accuracy, computational efficiency, and robustness when applied to multimodal stress detection on the WESAD and CogLoad datasets.
The default settings provide a reliable baseline, which allows us to assume that the performance change is solely initiated by the hyperparameter taken into consideration. Each hyperparameter is tested through a valid range of values, discussed in the search space section. 

}

{\color{black}
\section{Experimental Setup}
\subsection{Experimental Settings}

\noindent{\textbf{Hardware Platform:} All experiments were conducted on an 8GB VRAM NVIDIA GeForce RTX 2070 GPU-enabled laptop, which is supported by Microsoft Windows 10, 
and equipped with CUDA 12.7 and NVIDIA-SMI driver version 566.36. 
The GPU consumes power from 3W (idleness) to 41W (peak load). The system includes an Intel-based processor and 16GB of RAM, which are sufficient to conduct the computations for training and testing deep learning models.

\noindent{\textbf{Software Frameworks:}} We run our software on Python 3.9 and implement deep learning models using GitHub codes \cite{SMARTlab-Purdue} based on the PyTorch 1.8 framework. CUDA is used to enable GPU acceleration for parallel processing.

\begin{table}[t]
\caption{Specifications of Experimental Platform}
\centering
\resizebox{\columnwidth}{!}
{
\begin{tabular}{|p{3cm}|p{5cm}|}
\hline
\multicolumn{2}{|c|}{\textbf{Laptop Platform}} \\
\hline
GPU & 2304-core NVIDIA GeForce RTX 2070, 8GB VRAM, CUDA 12.7, NVIDIA-SMI 566.36 \\
\hline
CPU & Intel-based processor \\
\hline
Memory & 16GB RAM \\
\hline
Operating System & Microsoft Windows 10 
\\ \hline
Power & 3W (idleness) to 41W (peak load) \\
\hline
\multicolumn{2}{|c|}{\textbf{Software Frameworks}} \\
\hline
Programming Language & Python 3.9 \\
\hline
Deep Learning Framework & PyTorch 1.8 
\\ \hline
\end{tabular}
}
\label{tab:specs}
\end{table}

\subsection{Datasets}
The study makes use of two datasets: Wearable Stress and Affect Detection (WESAD) \cite{wesad} and Cognitive Load (CogLoad) \cite{cogload}.

\noindent{\textbf{WESAD dataset:}} 
It is a popular open-source multimodal stress and emotion recognition dataset within the fields of physiological and affective computing. The dataset comprises multimodal physiological sensor data recorded from 15 subjects, with each subject experiencing a controlled experiment for inducing various affective states, such as neutral, stress, amusement, and baseline states. Data were collected with two wearable sensors: RespiBAN, a chest-worn sensor that delivers Galvanic Skin Response (GSR), Respiration (RESP), Electrocardiogram (ECG), and Electromyogram (EMG); and Empatica E4, a wrist-worn sensor that records wrist-based GSR and Blood Volume Pulse (BVP).


To facilitate model input, these signals were segmented into fixed-size windows (700 samples for chest signals and 64 or 4 samples for wrist signals), with each segment reshaped to structured input formats suitable for model training. Only segments with consistent labels across the entire window were retained, and segments with ambiguous labeling (i.e., multiple labels within a window) were discarded. Furthermore, to focus on meaningful emotional states, segments corresponding to neutral labels were filtered out.

\noindent{\textbf{CogLoad dataset:}} 
The dataset \cite{cogload} is the first dataset that enables the analysis of the cognitive load associated with six different tasks based on an individual's physiological responses and personality traits. It is an open-source dataset that is ready-to-use in \texttt{.pkl} file format obtained from \cite{husformer}. This dataset contains physiological signals (EEG, GSR, BVP), collected from 23 participants using a Microsoft Band. Each participant completed a rest session and six dual-task tasks, all of which imposed cognitive load. The first task of the dual-task was randomly selected from standard psycho-physiological tasks from the work of Haapalainen et al. \cite{cogtasks}. 
%

\subsection{Evaluation}


Following the original Husformer evaluation practice \cite{SMARTlab-Purdue}, we adopted a modified 10-fold cross-validation (CV) strategy. 
This approach enables robust model validation by systematically rotating training, validation, and testing splits across the data.
Given $N$ samples, the data are partitioned into 10 approximately equal-sized subsets or \emph{folds}. For each fold $i \in \{0, 1, \dots, 9\}$, the following procedure is executed:

\begin{itemize}
    \item \textbf{Validation Set:} The $i$-th fold, corresponding to indices from $i \cdot \frac{N}{10}$ to $(i+1) \cdot \frac{N}{10}$, is designated as the validation set.
    \item \textbf{Test Set:} The subsequent $(i+1)$-th fold is used as the test set. For the last fold ($i = 9$), the test set wraps around and includes the first $\frac{N}{10}$ samples, preserving fold size consistency.
    \item \textbf{Training Set:} All remaining samples not included in the current validation or test sets are assigned to the training set.
\end{itemize}

This yields non-overlapping training, validation, and test sets for each of the 10 folds, with each sample participating in all three roles (training, validation, testing) across the full cross-validation cycle.

In the \texttt{WESAD()} function, this procedure is implemented using slicing operations on the shuffled index array. For each fold, the respective sample indices for training, validation, and testing are passed to the \texttt{pkl\_make()} function. This function constructs and serializes the fold-specific datasets into separate \texttt{.pkl} files.
Each \texttt{.pkl} file contains input data from six physiological modalities, along with corresponding labels and sample identifiers. This setup enables a consistent evaluation of the model’s performance across multiple independent folds, with the final performance metric averaged over all folds to reduce the risk of overfitting or sample-specific bias.

\begin{table*}[!t]
\centering
\caption{Layers configuration results on WESAD and CogLoad datasets (Default: $c\_layers=s\_layers=5$, $c\_heads=s\_heads=3$, $d_m=30$, $FFN=120$); The best results are shown in {\bf bold}, and the worst results are \underline{underlined})
}
\resizebox{\textwidth}{!}{%
\begin{tabular}{|c|c|c|c|c|c|c|c|c|c|}
\hline
\rowcolor{yellow!20}
\textbf{No} & \textbf{Cross-Modal Layers} & \textbf{Self-Attention Layers} & \textbf{Loss} & \textbf{MAE} & \textbf{Accuracy} & \textbf{F1} & \textbf{Train Time (h)} & \textbf{Memory (MB)} & \textbf{Params} \\
\hline
\multicolumn{10}{|c|}{\cellcolor{gray!10}\textbf{WESAD}} \\
\rowcolor{green!20}
\textbf{1 }& \textbf{1} & \textbf{1} & \textbf{0.2661} & \textbf{0.1266}& \textbf{0.8951} & \textbf{0.8955} & \textbf{0.45} & \textbf{653.68}& \textbf{81,910} \\
2 & 2 & 2 & 0.4868 & 0.2320 & 0.8240 & 0.8165 & 0.75 & 1157.89 & 160,240 \\
3 & 3 & 3 & 0.4682 & 0.2232 & 0.8295 & 0.8125 & 1.09 & 1652.86 & 238,570 \\
4 & 4 & 4 & 0.5166 & 0.2549 & 0.8053 & 0.8015 & 1.36 & 2155.12 & 316,900 \\
\underline{5} & \underline{5} & \underline{5} & \underline{0.5862} & \underline{0.2715} & \underline{0.7766} & \underline{0.7787} & \underline{1.74} & \underline{2652.86} & \underline{395,230} \\
\hline
\multicolumn{10}{|c|}{\cellcolor{gray!10}\textbf{CogLoad}} \\
\rowcolor{green!20}
\textbf{1} & \textbf{1} & \textbf{1} & \textbf{0.0413} & \textbf{0.0217}& \textbf{0.9643} & \textbf{0.9639} & \textbf{0.31} & \textbf{420.51} & \textbf{57,398} \\
2 & 2 & 2 & 0.0645 & 0.0353 & 0.9457 & 0.9509 & 0.56 & 780.12 & 113,348 \\
3 & 3 & 3 & 0.0633 & 0.0292 & 0.9460 & 0.9410 & 0.79 & 1132.78 & 169,298 \\
4 & 4 & 4 & 0.0684 & 0.0365 & 0.9436 & 0.9382 & 1.06 & 1485.34 & 225,248 \\
\underline{5} & \underline{5} & \underline{5} & \underline{0.0695} & \underline{0.0375} & \underline{0.9427} & \underline{0.9378} & \underline{1.41} & \underline{1837.90} & \underline{281,198} \\
\hline
\end{tabular}
}
\label{tab:combined_layer_priority}
\end{table*}

\subsection{Quantitative Metrics}
Our evaluation metrics from classification and computational perspectives are as follows: average multi-class accuracy (Acc), average multi-class F1 score (F1), cross-entropy loss (CE), mean absolute error (MAE), training duration (in hours), memory usage (in MB), and total trainable parameters. 
Let $n$ be the number of classes. The average multi-class accuracy is defined as: 
\begin{equation}
Acc = \frac{1}{n} \sum_{i=1}^{n} accuracy_i
\end{equation}
where $accuracy_i$ is the binary accuracy for class $i$, given by: 
\begin{equation}
accuracy_i = \frac{TP_i + TN_i}{length_i}
\end{equation} Here, $TP_i$ and $TN_i$ denote the $true$ $positive$ and $true$ $negative$ predictions for class $i$, respectively, and $length_i$ refers to the number of samples in that class.

The macro-averaged F1 score is computed using: 
\begin{equation}
F1 = \frac{1}{n} \sum_{i=1}^{n} f1_i
\end{equation}
where $f1_i$ is the F1 score for class $i$, defined as the harmonic mean between $precision_i$ and $recall_i$, precision and recall for class $i$:
\begin{equation}
f1_i = \frac{2 \cdot precision_i \cdot recall_i}{precision_i + recall_i}
\end{equation}
Precision and recall are given by:
\begin{equation}
precision_i = \frac{TP_i}{TP_i + FP_i}, \quad recall_i = \frac{TP_i}{TP_i + FN_i}
\end{equation}
where $FP_i$ and $FN_i$ are the number of $false$ $positive$ and $false$ $negative$ predictions for class $i$, respectively. 

In terms of loss-based evaluation, the cross-entropy loss $CE$ is additionally included, defined over all $N$ samples and $n$ classes as:
\begin{equation}
CE = - \frac{1}{N} \sum_{j=1}^{N} \sum_{i=1}^{n} y_{ji} \log(\hat{y}_{ji})
\end{equation}
where $y_{ji}$ is the ground truth label and $\hat{y}_{ji}$ is the predicted probability for class $i$ on sample $j$.

We also report the mean absolute error $MAE$, a general-purpose metric, indicating average prediction error magnitude:
\begin{equation}
MAE = \frac{1}{N} \sum_{j=1}^{N} \left| y_j - \hat{y}_j \right|
\end{equation}
where $y_j$ and $\hat{y}_j$ are the ground truth label and the predicted probability for the sample $j$, respectively.

To supplement the classification metrics, we also report the computational cost. Specifically, we provide the total training time $T_{\text{train}}$, defined as:
\begin{equation}
T_{\text{train}} = \sum_{e=1}^{E} t_e
\end{equation}
where $t_e$ indicates the elapsed time for the epoch $e$, and $E$ is the total number of training epochs.

Memory efficiency is measured through maximum memory consumption $M_{\text{peak}}$ at training time $t$:
\begin{equation}
M_{\text{peak}} = \max_{t \in [0, T_{\text{train}}]} memory(t)
\end{equation} 

Additionally, we count the number of trainable parameters $N_{\text{params}}$ to account for model complexity:
\begin{equation}
N_{\text{params}} = \sum_{l=1}^{L} P_l
\end{equation}
where $P_l$ refers to the number of parameters in the layer $l$, and $L$ is the total number of layers.
}

{\color{black}
\section{Results and Analysis}





\subsection{Quantitative Assessment of Key Hyperparameters}

Based on all of the experiments conducted in this study, we organize an evidence-based order for modifying the Transformer architecture: the order is to change from the attention layers ($L$) and heads ($H$), the model dimension ($d_m$), and finally to the FFN hidden size ($FFN$). 
This order is driven by empirical performance and cost.

\subsubsection{Layer Effects}
%
Table~\ref{tab:combined_layer_priority} presents the results of varying the number of cross-modal ($L_{cm}$) and self-attention ($L_{sa}$) layers  from 1 to 5, with other configurations fixed at the default. For both datasets, surprisingly increasing the number of layers leads to consistent performance degradation. 
On the WESAD dataset, the configuration with a single layer in both the cross-modal and self-attention blocks achieves the highest accuracy (89.51\%) and F1 score (89.55\%), requiring significantly less computation and enabling faster training. 
As the depth increases, the accuracy steadily drops to 77.66\% with the five-layer architecture. A similar trend is observed on the CogLoad dataset, where the one-layer model attains the accuracy of 96.43\%, outperforming those of deeper models, in which the five-layer drops to 94.27\%.

\begin{figure}
    \centering
    \includegraphics[width=\linewidth]{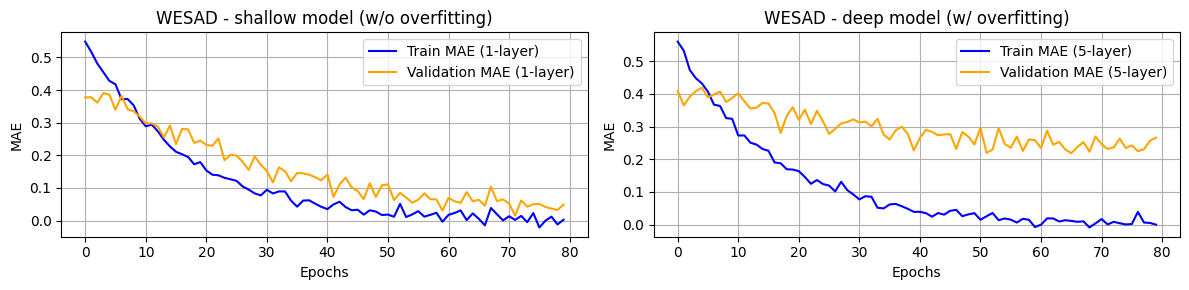}
    
    \vspace{0.5cm}
    
    \includegraphics[width=\linewidth]{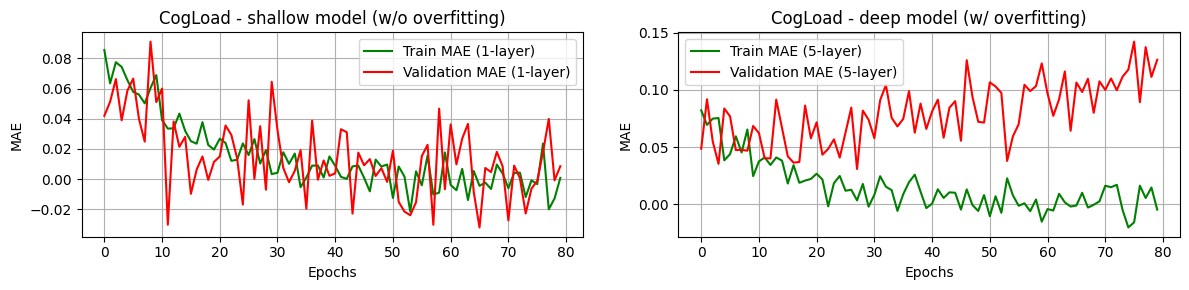}
    
    \caption{Comparison of Training and Validation Errors during Training between 1-layer (left) and 5-layers (right) on WESAD (top) and CogLoad (bottom) datasets
    }
    \label{fig:layers-comparison}
\end{figure}

As shown in Figure \ref{fig:layers-comparison}, we compare the training and validation error patterns on WESAD and CogLoad datasets. 
The 5-layer Transformers demonstrate the patterns of overfitting trend, a phenomenon commonly associated with increasing model depth in DL architectures. This tendency is particularly evident when applied to small and structured datasets. 
The phenomenon of diminishing returns indicates that the addition of more layers primarily amplifies noise in the data rather than contributing to the extraction of meaningful hierarchical features and latent patterns~\cite{zhang2016understanding}~\cite{brigato2021close}. 
As a result, the gap between training and validation performance widens; while training error decreases, validation error stagnates on WESAD or rises on CogLoad in later epochs. 
%

\subsubsection{Head Effects}
%
Table \ref{tab:combined_heads} explores the performance results of changing the number of attention heads in both Transformer modules, while keeping other parameters fixed. The results indicate that a single attention head in each module yields the best performance across both datasets. On WESAD, increasing the number of heads from one to three results in a noticeable drop in accuracy, from 81.96\% to 77.66\%, accompanied by a significant increase in memory usage. Similarly, the CogLoad exhibits a decline in performance as the number of heads increases.

The observed performance degradation with additional heads is likely due to feature space fragmentation. With small hidden dimensions (e.g., $d_m = 30$), multi-head attention divides the limited representation into even smaller subspaces, weakening the expressiveness of each head. Since both datasets consist of relatively low-dimensional and temporally stable signals, the complexity added by multiple attention heads becomes redundant and introduces variance that harms generalization~\cite{voita2019analyzing}~\cite{bhojanapalli2020lowrank}~\cite{hrycej2022heads}.

\begin{table*}[!t]
\centering
\caption{Heads configuration results on WESAD and CogLoad datasets (Default: $c\_layers=s\_layers=5$, $c\_heads=s\_heads=3$, $d_m=30$, $FFN=120$). 
The best results are shown in {\bf bold}, and the worst results are \underline{underlined}.
}
\footnotesize
\begin{tabular}{|c|c|c|c|c|c|c|c|c|c|}
\hline
\rowcolor{yellow!20}
\textbf{No} & \textbf{Heads ($H_{cm}$)} & \textbf{Heads ($H_{sa}$)} & \textbf{Loss} & \textbf{MAE} & \textbf{Accuracy} & \textbf{F1} & \textbf{Train Time (h)} & \textbf{Memory (MB)} & \textbf{Params} \\
\hline
\multicolumn{10}{|c|}{\cellcolor{gray!10}\textbf{WESAD}} \\
\rowcolor{green!20}
\textbf{1} & \textbf{1} & \textbf{1} & \textbf{0.4629} & \textbf{0.2434} & \textbf{0.8196} & \textbf{0.8116} & \textbf{1.42} & \textbf{1583.21} & \textbf{395,230} \\
2 & 2 & 2 & 0.5552 & 0.2830 & 0.7820 & 0.7796 & 1.51 & 2091.66 & 395,230 \\
\underline{3} & \underline{3} & \underline{3} & \underline{0.5862} & \underline{0.2715} & \underline{0.7766} & \underline{0.7787} & \underline{1.74} & \underline{2652.86} & \underline{395,230} \\
\hline
\multicolumn{10}{|c|}{\cellcolor{gray!10}\textbf{CogLoad}} \\
\rowcolor{green!20}
\textbf{1} & \textbf{1} & \textbf{1} & \textbf{0.0532} & \textbf{0.0279} & \textbf{0.9542} & \textbf{0.9527} & \textbf{1.39} & \textbf{1432.47} & \textbf{281,198} \\
2 & 2 & 2 & 0.0668 & 0.0381 & \underline{0.9360} & \underline{0.9348} & 1.37 & 1635.81 & 281,198 \\
\underline{3} & \underline{3} & \underline{3} & \underline{0.0695} & \underline{0.0375} & 0.9427 & 0.9378 & \underline{1.41} & \underline{1837.90} & \underline{281,198} \\
\hline
\end{tabular}
\label{tab:combined_heads}
\end{table*}

\begin{table*}[!t]
\centering
\caption{Dimension size configuration results on WESAD and CogLoad datasets (Default: $c\_layers=s\_layers=5$, $c\_heads=s\_heads=3$, $d_m=30$, $FFN=120$); The best results are shown in {\bf bold}, and the worst results are \underline{underlined})
}
\footnotesize
\begin{tabular}{|c|c|c|c|c|c|c|c|c|c|}
\hline
\rowcolor{yellow!20}
\textbf{No} & \textbf{ Dimension Size ($d_m$)} & \textbf{Loss} & \textbf{MAE} & \textbf{Accuracy} & \textbf{F1} & \textbf{Train Time (h)} & \textbf{Memory (MB)} & \textbf{Params} \\
\hline
\multicolumn{9}{|c|}{\cellcolor{gray!10}\textbf{WESAD}} \\
\underline{1} & \underline{9} & \underline{0.6780} & \underline{0.3604} & \underline{0.7373} & \underline{0.7152} & \textbf{1.52} & \textbf{1993.34} & \textbf{39,154} \\
\rowcolor{green!20}
\textbf{2} & \textbf{18} & \textbf{0.4896} & \textbf{0.2597} & \textbf{0.7932} & \textbf{0.7932} & 1.61 & 2289.62 & 146,290 \\
3 & 30 & 0.5111 & 0.2658 & 0.7767 & 0.7787 & \underline{1.74} & \underline{2652.86} & \underline{395,230} \\
\hline
\multicolumn{9}{|c|}{\cellcolor{gray!10}\textbf{CogLoad}} \\
\underline{1} & \underline{9} & \underline{0.0863} & \underline{0.0526} & \underline{0.9183} & \underline{0.9114} & \textbf{1.38} & \textbf{782.33} & \textbf{103,706} \\
2 & 18 & 0.0595 & 0.0361 & 0.9351 & 0.9288 & \underline{1.46} & 1315.42 & 189,698 \\
\rowcolor{green!20}
\textbf{3} & \textbf{30} & \textbf{0.0695} & \textbf{0.0375} & \textbf{0.9427} & \textbf{0.9378} & 1.41 & \underline{1837.90} & \underline{281,198} \\
\hline
\end{tabular}
\label{tab:dm_combined_priority}
\end{table*}

\begin{table*}[!t]
\centering
\caption{FFN dimension size configuration results on WESAD and CogLoad datasets (Default: $c\_layers=s\_layers=5$, $c\_heads=s\_heads=3$, $d_m=30$, $FFN=120$); The best results are shown in {\bf bold}, and the worst results are \underline{underlined})
}
\footnotesize
\begin{tabular}{|c|c|c|c|c|c|c|c|c|}
\hline
\rowcolor{yellow!20}
\textbf{No} & \textbf{~~~~~$FFN$ Size~~~~~} & \textbf{Loss} & \textbf{MAE} & \textbf{Accuracy} & \textbf{F1} & \textbf{Train Time (h)} & \textbf{Memory (MB)} & \textbf{Params} \\
\hline
\multicolumn{9}{|c|}{\cellcolor{gray!10}\textbf{WESAD}} \\
\rowcolor{green!20}
\textbf{1} & \textbf{30} & \textbf{0.4404} & \textbf{0.2419} & \textbf{0.8110} & \textbf{0.8113} & \textbf{1.67} & \textbf{2461.20} & \textbf{203,080} \\
2 & 60 & 0.5046 & 0.2434 & 0.7979 & 0.7970 & 1.69 & 2525.71 & 267,130 \\
3 & 90 & \underline{0.5278} & 0.2612 & 0.7877 & 0.7858 & \underline{1.74} & 2598.04 & 331,180 \\
\underline{4} & \underline{120} & 0.5111 & \underline{0.2658} & \underline{0.7767} & \underline{0.7787} & \underline{1.74} & \underline{2652.86} & \underline{395,230} \\
\hline
\multicolumn{9}{|c|}{\cellcolor{gray!10}\textbf{CogLoad}} \\
\rowcolor{green!20}
\textbf{1} & \textbf{30} & \textbf{0.0512} & \textbf{0.0287} & \textbf{0.9553} & \textbf{0.9521} & \textbf{1.35} & \textbf{1457.32} & \textbf{187,698} \\
2 & 60 & 0.0568 & 0.0325 & 0.9496 & 0.9467 & 1.37 & 1713.54 & 235,448 \\
3 & 90 & \underline{0.0723} & 0.0363 & 0.9438 & \underline{0.9332} & \underline{1.45} & 1594.77 & 143,948 \\
\underline{4} & \underline{120} & 0.0695 & \underline{0.0375} & \underline{0.9427} & 0.9378 & 1.41 & \underline{1837.90} & \underline{281,198} \\
\hline
\end{tabular}
\label{tab:ffn_combined_results}
\end{table*}

\subsubsection{Dimension Size Effects}
%
As shown in Table~\ref{tab:dm_combined_priority}, we evaluate the sensitivity of the model to variations in hidden dimension size $d_m$ by testing three configurations: $d_m = 9$, $18$, and $30$. 
On the WESAD dataset, the configuration with $d_m = 18$ achieves the best accuracy (79.32\%), outperforming both smaller and larger settings. While $d_m = 30$ yields a slight performance improvement on the CogLoad dataset (94.27\% accuracy), the gain over $d_m = 18$ remains marginal when considering the associated increase in computational and memory costs.

A small hidden dimension of $d_m$=9 leads to underparameterization, limiting the model's ability to capture complex patterns and resulting in insufficient representation capacity. 
Conversely, increasing the hidden dimension substantially inflates the number of parameters and memory requirements without proportionally improving the representation of sensor input~\cite{yao2017deepiot}~\cite{tang2020layerwise}. 
%
The trade-off shows that a moderate setting, such as $d_m = 18$ on WESAD and $d_m = 30$ on CogLoad dataset, respectively, provides the best balance between accuracy and efficiency, making it ideal for resource-constrained wearable sensing applications.  

\begin{table*}[!t]
\centering
\caption{Ablation Study Results on WESAD and CogLoad datasets
(Local Optimal Configurations: $c\_layers$ =$s\_layers=1$, $c\_heads$ =$s\_heads=1$, $d_m=18/30$, $FFN=30$); The best results are shown in {\bf bold}, and the worst results are \underline{underlined})
}
\resizebox{\textwidth}{!}{%
\begin{tabular}{|c|c|c|c|c|c|c|c|c|c|c|c|}
\hline
\rowcolor{yellow!20}
\textbf{Ablation Pair} & \textbf{L} & \textbf{H} & \textbf{$d_m$} & \textbf{FFN} & \textbf{Loss} & \textbf{MAE} & \textbf{Accuracy} & \textbf{F1} & \textbf{Train Time (h)} & \textbf{Memory Used (MB)} & \textbf{Params} \\
\hline
\multicolumn{12}{|c|}{\cellcolor{gray!10}\textbf{WESAD}} 
\\Default & \underline{5} & \underline{3} & \underline{30} & \underline{120} & 0.5111 & 0.2658 & 0.7767 & 0.7787 & \underline{1.74} & \underline{2652.86} & \underline{395,230} \\
L + H         & 1 & 1 & 30 & 120 & 0.4236 & 0.2135 & 0.8312 & 0.8235 & 0.39 & 408.92 & 81,910 \\
\rowcolor{green!20}
\textbf{L + $d_m$ }        & \textbf{1} & \textbf{3} & \textbf{18} & \textbf{120} & \textbf{0.2999} & \textbf{0.1486 }& \textbf{0.8841} & \textbf{0.8815} & \textbf{0.42} & \textbf{575.20} & \textbf{30,874} \\
\rowcolor{green!20}
\textbf{L + FFN }        & \textbf{1 }& \textbf{3} & \textbf{30} &  \textbf{30} & \textbf{0.3155 }& \textbf{0.1407} & \textbf{0.8815} & \textbf{0.8819} & \textbf{0.43} & \textbf{634.03 }& \textbf{43,480} \\
H + $d_m$         & 5 & 1 & 18 & 120 & 0.4767 & 0.2567 & 0.7917 & 0.7898 & 1.31 & 1178.01 & 146,290 \\
H + FFN         & 5 & 1 & 30 &  30 & 0.5020 & 0.2715 & 0.7957 & 0.7869 & 1.37 & 1368.12 & 203,080 \\
$d_m$ + FFN         & 5 & 3 & 18 &  30 & 0.3546 & 0.1649 & 0.8632 & 0.8644 & 1.59 & 2166.27 & 76,360 \\
L + H + $d_m$     & 1 & 1 & 18 & 120 & 0.3855 & 0.2096 & 0.8446 & 0.8460 & 0.35 & 304.45 & 30,874 \\
L + H + FFN     & 1 & 1 & 30 &  30 & 0.4021 & 0.2114 & 0.8299 & 0.8313 & 0.38 & 368.24 & 43,480 \\
L + $d_m$ + FFN     & 1 & 3 & 18 &  30 & 0.3859 & 0.2108 & 0.8479 & 0.8423 & 0.42 & 562.12 & 16,888 \\
H + $d_m$ + FFN     & \underline{5} & \underline{1} & \underline{18} &  \underline{30} & \underline{0.6260} & \underline{0.3090} & \underline{0.7737} & \underline{0.7566} & 1.30 & 1058.63 & 76,360 \\
L + H + $d_m$ + FFN & 1 & 1 & 18 &  30 & 0.4365 & 0.2289 & 0.8334 & 0.8235 & 0.37 & 287.53 & 16,888 \\
\hline
\multicolumn{12}{|c|}{\cellcolor{gray!10}\textbf{CogLoad}} 
\\ Default & \underline{5} & \underline{3} & \underline{30} & \underline{120} & 0.4412 & 0.2145 & 0.8657 & 0.8681 & \underline{1.92} & \underline{2798.77} & \underline{281,198} \\
L + H & 1 & 1 & 30 & 120 & 0.3711 & 0.1710 & 0.8920 & 0.8876 & 0.49 & 422.66 & 81,910 \\
\rowcolor{green!20}
\textbf{L + $d_m$} & \textbf{1} & \textbf{3} & \textbf{18} & \textbf{120} & \textbf{0.2634} & \textbf{0.1203} & \textbf{0.9245 }& \textbf{0.9229} & \textbf{0.51} & \textbf{598.45} & \textbf{30,874} \\
\rowcolor{green!20}
\textbf{L + FFN} & \textbf{1} & \textbf{3} & \textbf{30} & \textbf{30} & \textbf{0.2402} & \textbf{0.1115} & \textbf{0.9261} & \textbf{0.9273} & \textbf{0.52} & \textbf{644.81} & \textbf{43,480} \\
H + $d_m$ & 5 & 1 & 18 & 120 & 0.3980 & 0.1944 & 0.8788 & 0.8740 & 1.41 & 1213.22 & 146,290 \\
H + FFN & 5 & 1 & 30 & 30 & 0.4267 & 0.2022 & 0.8716 & 0.8665 & 1.48 & 1391.00 & 203,080 \\
$d_m$ + FFN & 5 & 3 & 18 & 30 & 0.2915 & 0.1333 & 0.9130 & 0.9116 & 1.66 & 2234.20 & 76,360 \\
L + H + $d_m$ & 1 & 1 & 18 & 120 & 0.3106 & 0.1581 & 0.9018 & 0.8993 & 0.46 & 318.90 & 30,874 \\
L + H + FFN & 1 & 1 & 30 & 30 & 0.3333 & 0.1634 & 0.8955 & 0.8920 & 0.48 & 377.81 & 43,480 \\
L + $d_m$ + FFN & 1 & 3 & 18 & 30 & 0.3120 & 0.1558 & 0.9040 & 0.9025 & 0.52 & 574.33 & 16,888 \\
H + $d_m$ + FFN & \underline{5} & \underline{1} & \underline{18} & \underline{30} & \underline{0.4925} & \underline{0.2271} & \underline{0.8571} & \underline{0.8532} & 1.39 & 1090.80 & 76,360 \\
L + H + $d_m$ + FFN & 1 & 1 & 18 & 30 & 0.3548 & 0.1767 & 0.8874 & 0.8855 & 0.45 & 306.02 & 16,888 \\
\hline
\end{tabular}
}
\label{tab:ablation_combined}
\end{table*}

\subsubsection{FFN Size Effects}
%
Table \ref{tab:ffn_combined_results} investigates the influence of the FFN dimension size on model performance. Configurations with FFN sizes of 30, 60, 90, and 120 are evaluated. The results indicate that smaller FFN sizes yield performance that is equal to or better than larger configurations. Specifically, on the CogLoad dataset, an FFN size of 30 achieves the highest accuracy and F1 score while consuming the least memory. For WESAD, an FFN size of 30 also outperforms larger alternatives in terms of both accuracy and efficiency.

These findings suggest that large FFN layers may introduce overfitting and unnecessary computation. In Transformer architectures, FFN layers serve as position-wise transformations that expand and project features. In scenarios with relatively simple temporal structures, as seen in physiological data, excessive FFN capacity may learn noise instead of meaningful signal features. A smaller FFN not only helps in regularization but also reduces model latency and parameter count \cite{hassani2021compact} \cite{pires2023onewide}.

\subsection{Ablation Study}

To further understand the impact of each architectural component, an ablation study was conducted, as shown in Table \ref{tab:ablation_combined}. We analyze the combinations of local optimal configurations of the number of layers (L), attention heads (H), model dimension ($d_m$), and FFN size.

On WESAD, the combination of 1-layer and $d_m = 18$ (L + $d_m$) results in a remarkable accuracy of 88.41\%, with only 30K parameters—less than one-tenth of that of the full model. 
Additionally, the combination of 1-layer and $FFN = 30$ (L + FFN) also results in a remarkable accuracy of 88.15\%, with only 43K parameters. 
Similarly, for CogLoad, (L + $d_m$) and (L + FFN) configurations achieve strong performance with minimal resource requirements.  
The best configuration (L + FFN) achieves 92.61\% accuracy with significantly reduced training time and memory usage, almost similar but outperforming the accuracy of 92.45\% in (L + $d_m$). 
Based on the ablation study, (L + $d_m$) reveals a higher contribution to performance across two datasets. On WESAD (L + $d_m$) parametrization yields the best result, outperforming (L + FFN) with the smallest number of 30K parameters; On CogLoad, the second best, but with a margin of only 0.173\%. 

These results highlight that shallow architectures, when combined with optimized low-dimensional embeddings and compact FFN layers, are sufficient for learning robust temporal and cross-modal representations. Furthermore, the minimal importance of the number of attention heads emphasizes that for low-resolution sensor inputs, even a single-head mechanism can capture modality-specific attention effectively.
Across the ablation study of all locally-optimal configurations, our Efficient-Husformer demonstrates significant computational advantages. 
The most compact variant (e.g., L = 1, H = 1, $d_m$ = 18, FFN = 30) achieves high classification accuracy of 83.34\% and 88.74\% on WESAD and CogLoad, respectively, with fewer than 20K parameters and less than 400 MB memory usage. 
These models can be trained within one hour and are suitable for real-time deployment on on-device or edge devices, such as wearables and smartphones. 
%

\subsection{Summary of Results and Analysis} 


\noindent\textbf{Key Findings}:
1) Firstly, shallower architectures (one cross-modal layer and one self-attention layer) consistently outperformed deeper alternatives on both WESAD and CogLoad datasets, achieving the highest accuracy and lowest error, while being the most computationally frugal. The deeper models suffered from over-parameterization and incurred further computational costs.
2) Secondly, configurations with the least number of attention heads (one cross-modal layer and one self-attention head) achieved the best balance of performance and efficiency. Adding additional attention heads to the configurations increased computational expenses without improving accuracy.
%
%
3) Finally, when model depth and either $d\_m$ or FFN were tuned jointly, optimal configurations were found. The best configuration was with one encoder layer and $d\_m=18$, offering the optimal balance of learning capacity and generalization. The patterns appearing in CogLoad were similar, but with a higher tolerance for model complexity, resulting in the best of $d\_m=30$, suggesting potentially higher separability of data.

The findings demonstrate that the Efficient-Husformer architecture significantly advances the balance of accuracy, efficiency, and resource cost trade-offs for multi-class stress classification from physiological data. 
In particular, the optimization strategy with decoupling of the cross-modal Transformer and self-attention Transformer was shown to be effective. It allows mode-fusion abilities and classification functionalities to be tuned independently, enabling more interpretability and resource efficiency. 

}

}

{\color{black}
\section{Discussion and Future Work} 

%

\noindent\textbf{Progressive Transformer Scaling Across Diverse Datasets}: 
Although we propose a simple yet effective Efficient-Husformer model, achieving superior accuracy with a reduced memory footprint and training time, challenges regarding scalability and generalization in model design still remain. 
A potential direction is the adoption of multi-stage (progressive) model scaling for Transformer architectures across various domains, including Computer Vision~\cite{li2022automated}~\cite{li2024efficient}, Natural Language Processing~\cite{gu2021transformer_growth}~\cite{shen2022staged}, and Time Series data~\cite{zhou2023multi}~\cite{liang2024minusformer}.
Developing such multi-stage (progressive) models to accommodate diverse datasets will be the focus of our future work. 
}

{\color{black}
\section{Conclusion} 



We introduced Efficient-Husformer, a Transformer-based architecture designed for multimodal physiological stress detection to emphasize efficiency, 
building upon the Husformer backbone. 
Efficient-Husformer separates cross-modal and self-attention mechanisms, enabling independent configuration of architectural components such as the number of layers ($L$), attention heads ($H$), model dimension ($d_m$), and feed-forward network size ($FFN$). 
This design supports fine-grained optimization for physiological computing tasks while reducing computational overhead.

Through extensive hyperparameter optimization (HPO) and ablation studies on two benchmark datasets: WESAD and CogLoad, we demonstrated that Efficient-Husformer outperforms the original Husformer while maintaining compact and deployable memory footprints, 
with the best configurations achieving accuracies of 88.41\%  and 92.61\%  (improvements of 13.83\% and 6.98\%, compared to the original Husformer) on WESAD and CogLoad datasets, respectively. 
%
%
%
%
Importantly, these gains were achieved with a dramatic reduction in model size and training time, highlighting that high-performance Transformer models can be made lightweight and resource-efficient through targeted architectural tuning.

%

}
}

\bibliographystyle{ieeetr}
\bibliography{bibliography}

\newpage

 




\vfill

\end{document}